\documentclass[letterpaper, 10 pt, journal, twoside]{IEEEtran}
\ifCLASSINFOpdf
\else
\fi
\usepackage{graphics} % for pdf, bitmapped graphics files
\usepackage{epsfig} % for postscript graphics files
\usepackage{amsmath} % assumes amsmath package installed
\usepackage{amssymb}  % assumes amsmath package installed

\usepackage{algorithm}
\usepackage{algpseudocode}
\usepackage{graphicx}
\usepackage{float}
\usepackage{subcaption}
\usepackage{color}
\usepackage{multirow}
\usepackage{array}
\usepackage{caption}
\usepackage{tabularx}
\usepackage{booktabs}
\usepackage{amsmath}
\usepackage[dvipsnames]{xcolor}
\usepackage{dashrule}
\usepackage{tikz}
\usepackage{makecell}

\usepackage{hyperref}
\usepackage{siunitx}

\newcommand{\eg}{e.g., }
\newcommand{\ie}{i.e., }

\newcommand{\revision}[1]{\textcolor{black}{#1}}

\begin{document}

\title{Online Temporal Fusion for Vectorized Map Construction \\ in Mapless Autonomous Driving}

\author{Jiagang Chen$^{1}$, Liangliang Pan$^{2}$, Shunping Ji$^{1}$, Ji Zhao$^{2}$, and Zichao Zhang$^{2}$
\thanks{Manuscript received: August, 25, 2024; Revised January, 6, 2025; Accepted February, 4, 2025. %Use only for final RAL version
This paper was recommended for publication by Editor Ashis Banerjee upon evaluation of the Associate Editor and Reviewers' comments.
% This work was supported by (organizations/grants which supported the work.)
\textit{(Jiagang Chen and Liangliang Pan contributed equally to this work.) (Corresponding author: Zichao Zhang.)}} %Use only for final RAL version
\thanks{$^{1}$Jiagang Chen and Shunping Ji are with School of Remote Sensing and Information Engineering, Wuhan University, Wuhan 430079, China {(e-mail: chenjiagang2015@whu.edu.cn; jishunping@whu.edu.cn)}.}%
\thanks{$^{2}$Liangliang Pan, Ji Zhao, and Zichao Zhang are with Huixi Technology, China {(e-mail: ll.pan931204@gmail.com; zhaoji84@gmail.com; zhangzichao17@gmail.com)}.}%
\thanks{Digital Object Identifier (DOI): see top of this page.}
}

\markboth{IEEE Robotics and Automation Letters. Preprint Version. Accepted February, 2025}
{Chen \MakeLowercase{\textit{et al.}}: Online Temporal Fusion for Vectorized Map Construction in Mapless Autonomous Driving}

\maketitle

%%%%%%%%%%%%%%%%%%%%%%%%%%%%%%%%%%%%%%%%%%%%%%%%%%%%%%%%%%%%%%%%%%%%%%%%%%%%%%%%
\begin{abstract}

To reduce the reliance on high-definition (HD) maps, a growing trend in autonomous driving is leveraging onboard sensors to generate vectorized maps online.
However, current methods are mostly constrained by processing only single-frame inputs, which hampers their robustness and effectiveness in complex scenarios. 
To overcome this problem, we propose an online map construction system that exploits the long-term temporal information to build a consistent vectorized map.
First, the system efficiently fuses all historical road marking detections from an off-the-shelf network into a semantic voxel map, which is implemented using a hashing-based strategy to exploit the sparsity of road elements.
Then reliable voxels are found by examining the fused information and incrementally clustered into an instance-level representation of road markings.
Finally, the system incorporates domain knowledge to estimate the geometric and topological structures of roads, which can be directly consumed by the planning and control (PnC) module.
Through experiments conducted in complicated urban environments, we have demonstrated that the output of our system is more consistent and accurate than the network output by a large margin and can be effectively used in a closed-loop autonomous driving system.

\end{abstract}

% Keywords command
\providecommand{\keywords}[1] {
    \small	
    \textbf{\textit{index terms---}}
}

\hspace*{\fill}


\IEEEpeerreviewmaketitle

\section{Introduction}

\IEEEPARstart{M}{aps} play a crucial role in autonomous driving, serving as the foundational element for localization and PnC. In recent years, pre-produced HD maps have been the primary format utilized by autonomous vehicles. However, the production and maintenance of HD maps typically rely on professional mobile mapping systems equipped with multiple sensors, including LiDARs, global navigation satellite system (GNSS), inertial navigation system (INS), and cameras. This reliance makes the process both costly and time-consuming. 

To reduce the dependence on offline HD maps, mapless solutions have been developed that leverages on-board sensors to construct online maps in real time, and this paradigm shift also eliminates the need for localization in pre-built maps.
Representative methods include HDMapNet~\cite{li2022hdmapnet} and MapTR~\cite{liao2022maptr}, which construct maps for each individual frame from a Bird’s Eye View (BEV) using surround-view images.

However, relying on single-frame inputs makes these methods susceptible to occlusions and variable lighting.
The lack of temporal consistency between maps at different timestamps creates significant challenges for PnC modules.
Researchers have thus explored different approaches to improve the temporal consistency.
Temporal fusion using deep networks, such as StreamMapNet~\cite{yuan2024streammapnet} and MapTracker~\cite{chen2024maptracker}, uses memory buffers to retain previously encoded network features, and combines them with current data to improve map consistency. 
This, however, requires processing and storing a larger volume of network features with increased computational cost and inference time.
Tracking and fusing 3D detections from neural networks, such as MonoLaM~\cite{qiao2023online}, is more flexible and efficient, but depends on the temporal consistency of the detection results to ensure reliable tracking.
In addition, compared with mature HD maps, most existing online map construction methods lack the lane-level information, which are pivotal for PnC algorithms.

\begin{figure}[t]
\centering
\includegraphics[width=0.99\linewidth, trim={0 0.1cm 0 0},clip]{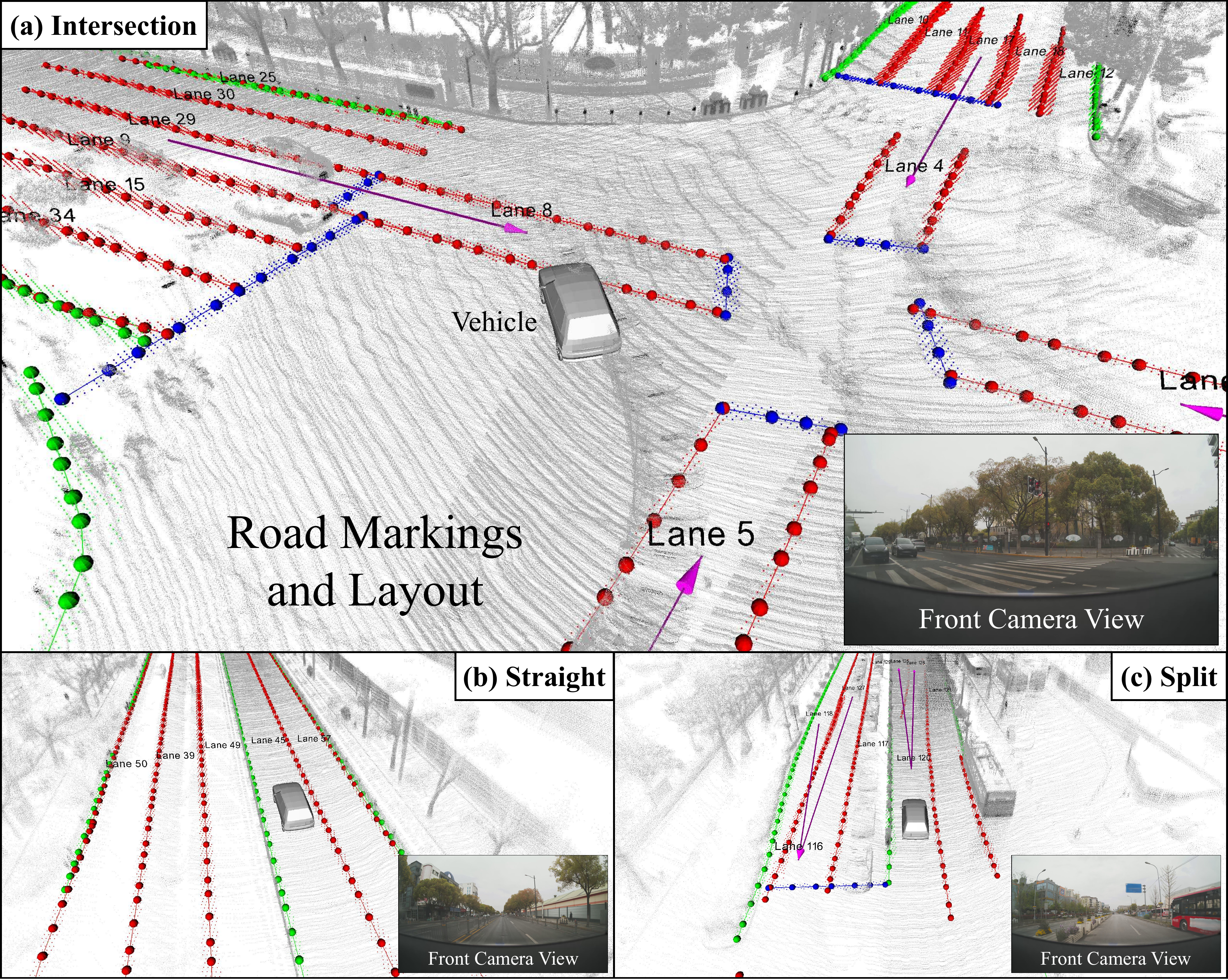} 
\caption{Results of the proposed system in an urban road test. \textbf{(a)-(c)} the road markings (\textcolor{blue}{stoplines}, \textcolor{Red}{lanelines}, and \textcolor{ForestGreen}{roadedges}) and layout (lanes and \textcolor[RGB]{209,6,207}{lane linkages}) generated by our method online in typical scenarios.
}
%including intersection, split, and straight scenes.}
\label{fig_teaser}
% \vspace{-10pt}
\end{figure}

In this work, we propose an online map construction system for mapless autonomous driving. The system takes road marking detections and vehicle poses as input and builds a detailed map around the vehicle in real-time.
The system is intended to be used in a closed-loop autonomous driving system, which requires the map to be consistent over time and contain lane-level information besides the vectorized road markings.
First, to maintain online efficiency and exploit long-term temporal information for consistency, we choose to fuse the output of detection networks instead the high-dimensional features, similar to \cite{qiao2023online}.
Differently, we adopt a sparse voxel map based on spatial hashing~\cite{niessner2013real} and fuse the detections over time into the voxels.
The voxel-based fusion has the advantage of requiring no explicit data association over time.
Moreover, stable road markings instances can be efficiently extracted from the voxels using properly designed filtering and clustering strategies.
In practice, our voxel-based approach is capable of generating consistent instances for different lane layouts (see Fig.~\ref{fig_teaser}).
Second, we propose a general procedure, by incorporating the domain knowledge of common road layouts, to build lanes and linkages from the vectorized instances extracted from the voxels.
This goes one step beyond the common output (\ie vectorized road markings) in previous works.
The lane-level output can be directly used by PnC algorithms, and, in our experiments, is sufficient to support autonomous driving in complicated urban environments.
To summarize, the key contributions of this work are as follows:

\begin{itemize}
\item An online map construction system for HD-mapless autonomous driving. It can directly output lightweight geometrical and topological road structures to be used by PnC algorithms.
\item An effective voxel-based fusion method that can efficiently fuse noisy road marking detection into consistent and accurate vectorized instances.
\item Comprehensive experiments in complex real-world urban environments to validate the effectiveness and adaptability of the proposed system.
\end{itemize}

\section{Related Work}

\subsection{3D Road Marking Detection}

The most straightforward way to obtain 3D road markings is to project semantic information from the image plane to 3D space.
For example, RoadMap~\cite{qin2021light} projects semantic segmentation in the front-facing camera to the ground plane using inverse perspective mapping (IPM).
However, IPM assumes that the ground surface is planar and thus can degrade in conditions like uphill roads.
Additional hardware, such as LiDARs~\cite{bai2018deep} and stereo cameras~\cite{fan2018real}, can be used to get the actual height for accurate projection.

Alternatively, researchers have explored to directly predict 3D road markings from images.
Instead of detecting road markings from a single camera as in~\cite{garnett20193d, chen2022persformer}, the majority of the recent research formulates the detection task in the BEV perspective to fully utilize the information from surround-view cameras.
In~\cite{zhou2022cross}, features (from convolutional backbones) of surround-view images are aggregated into BEV space, and the segmentation of road elements is performed on the BEV features.
To further facilitate downstream tasks, the vectorized form (\eg polylines) of the road elements can be generated either in a separate post-process step~\cite{li2022hdmapnet} or directly from the network in an end-to-end fashion~\cite{liu2023vectormapnet, liao2022maptr}.
Although state-of-the-art (SoTA) networks can already achieve excellent detection performance, they mostly rely on images \textit{at one single time} and are still lacking in stability in real-world scenarios. 

\subsection{Road Mapping}

One common approach to mitigate the instability of single detection is to fuse the information from multiple times to build a persistent map.
This is mostly developed as an offline process.
For example,~\cite{ebrahimi2022high} utilizes Simultaneously Localization and Mapping (SLAM) techniques with multiple sensors for offline HD map construction.
To minimize the required sensors and thus cost, RoadMap~\cite{qin2021light} takes an vision-based approach and builds an offline map for visual localization.

In contrast, online road mapping has the potential of reducing the offline maintenance cost and attracts increasing research interest. Most works adopt well-developed SLAM techniques.
Road-SLAM~\cite{jeong2017road} obtains road marking points using IPM and segments and classifies the points into different types. The classified markings are used as landmarks for loop closures in a pose-graph based SLAM.~\cite{zhou2022visual} also adopts IPM and utilizes road markings to constrain pose-graph optimization, but the markings are further vectorized as polylines and different error sources are carefully modeled.
Other works explore to use 3D road markings directly.
~\cite{meier2018visual} extracts 3D path boundaries, parameterized as B{\'e}zier curves, using stereo cameras. Then both the camera pose and the 3D curves are estimated in an extended Kalman Filter.~\cite{qiao2023online} proposes a SLAM system that uses 3D lanelines, modeled as Catmull-Rom splines, as landmarks. The system uses the 3D laneline detection from a deep neural network~\cite{chen2022persformer} as observation, and, combined with odometry poses, estimates the camera poses and spline parameters alternately.
Our work falls in the category of online road mapping but differs in several aspects. 
First, our approach does not require explicit data association as in~\cite{meier2018visual, qiao2023online}, thanks to the voxel-based fusion strategy.
Second, compared with other voxel-based approaches~\cite{qin2021light, jeong2017road}, we are able to generate vectorized road markings.
Finally, the proposed system focuses on generating directly usable map for downstream modules in real-world autonomous driving systems, while the aforementioned methods mostly serve the goal of localization.

 More recently, explicit temporal fusion has been introduced in detection networks directly to improve the stability.
 These methods, such as StreamMapNet~\cite{yuan2024streammapnet} and MapTracker~\cite{chen2024maptracker}, typically maintain a window of past network features to be fused with the current one.
 This incurs non-negligible computational cost and thus limits the span of temporal information that can be made use of.

\begin{figure*}[ht]
\centering
\includegraphics[width=0.99\linewidth, trim={0 0.1cm 0 0},clip]{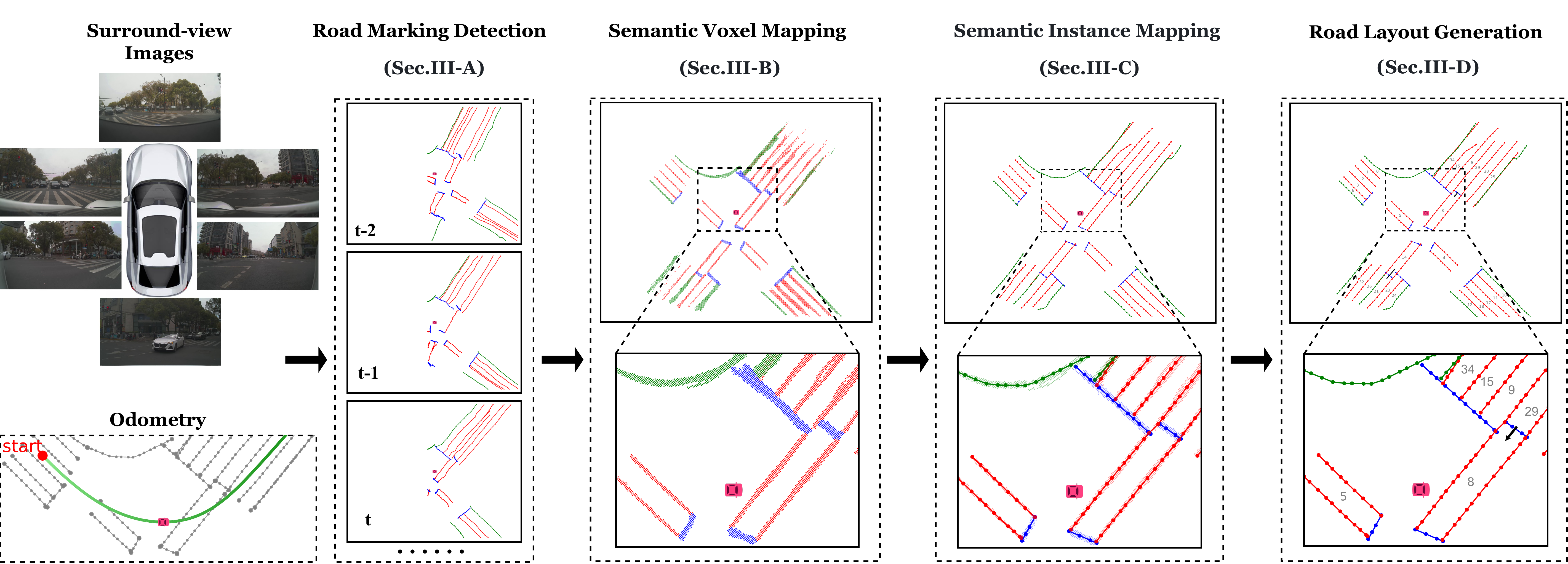} 
\caption{
Illustration of the proposed system. Utilizing surround-view cameras and odometry poses as input, the system outputs vectorized road markings (\textcolor{Red}{lanelines}, \textcolor{ForestGreen}{roadedges} and \textcolor{blue}{stoplines}) and road layout (lanes and lane linkages) for PnC modules.
}
% \vspace{-10pt}
\label{fig_framework}
\end{figure*}
\section{Methodology}

The proposed system, as llustrated in Fig.~\ref{fig_framework}, generates detailed road layout from the image streams of surround-view cameras and odometry poses.
The road information, including lanes and lane linkages, can be used by PnC algorithms directly.
In brief, the proposed approach systematically fuses the semantic information at different times to achieve accurate and temporally consistent result.
First, road markings are detected from the images at every frame (\ie multiple images at a single time) using a SoTA network, and the raw detection is filtered and transformed for later fusion (Sec.~\ref{sec_detect_preproc}).
Second, the preprocessed road markings are fused in to an efficient voxel-based representation.
Both the geometric and instance information are maintained and updated incrementally at every frame (Sec.~\ref{sec_semantic_vox_mapping}).
Then, the voxels are incrementally clustered and vectorized into instances, which correspond to actual lanelines, roadedges and stoplines (Sec.~\ref{sim_label}).
Finally, the road layout is generated from the instances by making use of the prior knowledge about common road structure (Sec.~\ref{sec_road_layout}).
As the PnC module only requires local road information rather than a global map, our system maintains the fused map within a fixed region centered at the vehicle.

\subsection{Road Marking Detection and Preprocessing}
\label{sec_detect_preproc}

Following standard practice, we use a deep neural network to detect the 3D road markings around the vehicle. The only assumption is that the network outputs the detected markings as polylines, which makes our method compatible with a wide range of networks (\eg~\cite{garnett20193d, chen2022persformer}). We use MapTR~\cite{liao2022maptr} in our system, as it is one of the SoTA approaches.
Concretely, each detected road marking $s_i \in \mathcal{S}$ is represented as ordered 3D points $\{\mathbf{p}^{\texttt{b}}_j\}_{j=1}^{M_i}$ with a semantic label and confidence, namely
\begin{align}
\mathcal{S} = \{s_i\}_{i=1}^{N}, \; s_i = (\{\mathbf{p}^{\texttt{b}}_j\}_{j=1}^{M_i},\:c_i,\:l_i),
\label{eq_network_detection}
\end{align}
where $l_i \in \mathcal{L} = \{\text{laneline}, \text{roadedge}, \text{stopline}\}$ is the road marking type and $c_i$ the confidence. Note that the 3D points are in the body frame \texttt{b} of the vehicle.

Due to imperfect network detection, we first discover and reject likely erroneous road markings in $\mathcal{S}$ according to several cues.
First, a detected marking $s_i$ is discarded if its confidence $c_i$ is lower than a certain threshold. Second, the angles between consecutive segments of the polyline $\{\mathbf{p}^{\texttt{b}}_j\}_{j=1}^{M}$ are computed, and zigzag polylines are rejected accordingly. 
The related thresholds are tuned manually.

The filtered road markings are then transformed to a predefined reference frame $\texttt{g}$ to be fused into a persistent representation.
Specifically, the points of the polyline is transformed as $\mathbf{p}^{\texttt{g}} = \mathbf{R}  \mathbf{p}^{\texttt{b}} + \mathbf{t}$, where $\mathbf{R}$ and $\mathbf{t}$ are the rotation and translation of the vehicle's body frame with respect to the reference frame.
For the simplicity of presentation, we use $\mathcal{S}$ and $s_i$ for the filtered and transformed road markings in the rest of the paper unless stated otherwise.

\subsection{Semantic Voxel Mapping} 
\label{sec_semantic_vox_mapping}

The filtered road markings at a single time can still contain considerable noise, and there is no enforcement of temporal consistency (see Fig.~\ref{fig_framework}, second column). Therefore, we fuse the detection at different times to obtain more stable geometric and semantic information, \ie a semantic map.
Moreover, the fusion is desired to be \textit{incremental} for online systems like ours, as the semantic map needs to be updated every time when new detection $\mathcal{S}^{(t)}$ becomes available.

\subsubsection{Voxel Map for Road Markings}
\begin{figure}[t]
\centering
\includegraphics[width=0.99\linewidth]{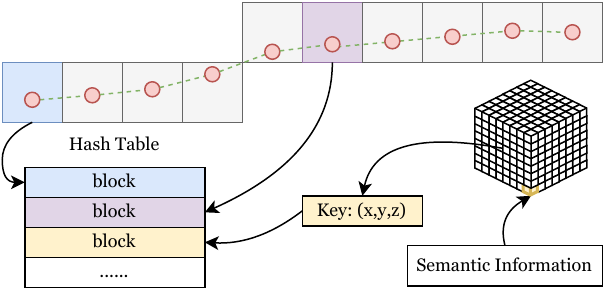} 
\caption{The voxel hashing data structure. The semantic voxel map is maintained in a hash table format and composed of independent blocks indexed by their spatial coordinates. Each block comprises an $8^3$ voxel grid and stores the voxels as a array. A voxel can thus be uniquely identified by the key of the block it belongs to and its index in the block.}
% \vspace{-10pt}
\label{fig_voxel_map}
\end{figure}

We use a dynamically sized voxel map based on the voxel hashing method~\cite{niessner2013real} to efficiently fuse the detected road markings at different times.
As illustrated in Fig.~\ref{fig_voxel_map}, the voxel map is composed of independent blocks indexed by their spatial coordinates. A hash table maintains the mapping between block positions and their memory locations, enabling constant-time $\mathcal{O}(1)$ operations for insertions and look-ups under typical conditions\footnote{Hash collision is resolved, at a higher time complexity, using the chaining method, where each bucket stores a linked list of the entries with the same hash code.}.
It is worth noting that the method efficiently addresses map growth by only allocating blocks for valid observation. This makes it especially suitable for representing road markings, which are inherently sparse.
We refer the reader to~\cite{niessner2013real, Oleynikova17voxblox} for the details of the map representation.

In our approach, a semantic voxel map is created at the reference frame $\texttt{g}$, and a voxel in the semantic map stores both geometric and semantic information as
\begin{align}
v=(\mathbf{x}^{\texttt{g}}, \{n_o\}_{o=1}^{|\mathcal{L}|}),\;\mathcal{L}  = \{\text{laneline}, \text{roadedge}, \text{stopline}\}
\label{eq_voxel_2}
\end{align}
where $\mathbf{x}^{\texttt{g}}$ is the center of the voxel, and $n_o$ represents  the number of times the voxel
%(\ie the volume that it occupies)
is detected as a certain road marking type. $\mathbf{x}^{\texttt{g}}$ is calculated from the spatial coordinates of the block the voxel belongs to and its index in the block.

\subsubsection{Incremental Voxel Map Update}
\label{subsec_inc_vox_update}
When new observations $\mathcal{S}^{(t)}=\{s_i\}_{i=1}^{N^{(t)}}$ are available, each road marking $s_i$ is integrated into the semantic voxel map as follows.
First, we identify the locations of all possible voxels that overlap with $s_i$ by sampling the polyline  $\{\mathbf{p}_j^{\texttt{g}}\}_{j=1}^{M_i}$ according to a predefined voxel size.
Then we allocate new voxels in the semantic map for locations that have not been observed before.
Finally, we update the detection times of all voxels as ${n}_{o_{i}}^{(t)} \gets {n}_{o_{i}}^{(t-1)} + 1$, where $o_{i}$ selects the same road marking type as $s_i$.

To facilitate the clustering of the voxels in following step (Sec.~\ref{sim_label}), we additionally maintain the instance-level information regarding the observed voxels.
Conceptually, we maintain a co-observation matrix $\mathcal{A}$ that can be indexed by any two voxels in the semantic map, where
\begin{align}
n_{v_p, v_q} = \mathcal{A}(v_p, v_q) = \mathcal{A}(v_q, v_p)
\label{eq_co_obs_mat}
\end{align}
denotes the number of times $v_p$ and $v_q$ appear in the same road marking detection.
Intuitively, $n_{v_p, v_q}$ indicates the likeliness that the two voxels belong to the same physical road marking.
To update $\mathcal{A}$ with $s_i$, we simply increment the corresponding entries of all the pairs of the overlapping voxels by one. For better space complexity, the co-observation matrix is implemented as a hash table.

\subsubsection{Reliable Voxel Extraction}
The update process described above accumulates the information of all historical detection per voxel.
This allows us to select reliable voxels, which have accumulated sufficient observations to be determined as a certain type of road markings.
Specifically, considering a voxel as formulated in \eqref{eq_voxel_2}, let $n^{*} = \max(\{n_o\}_{o=1}^{|\mathcal{L}|})$ be the detection times of the most likely road marking type, \revision{the voxel is considered reliable only when $n^{*} > \alpha_{n}$, where $\alpha_{n}$ is heuristically tuned thresholds.}

After processing all the road markings in $\mathcal{S}^{(t)}$, we can now obtain a collection of voxels that become reliable due to the latest detection.
Out of all the voxels that are updated as in Sec.~\ref{subsec_inc_vox_update}, the reliable ones are denoted collectively as
\begin{align}
\mathcal{V}^{(t)} = \{v^{*}_{k}\}_{k=1}^{K}, \; v^{*}_k = (\mathbf{x}^{\texttt{g}}_k, n^{*}_k, l^{*}_k),
\label{eq_new_mature_vox}
\end{align}
where $l^{*}_k \in \mathcal{L}$ is the most likely road marking type, and $n^{*}_k$ the corresponding detection times.
$\mathcal{V}^{(t)}$ are used to update the instance-level information, as will be described below.

\subsection{Semantic Instance Mapping\label{sim_label}}

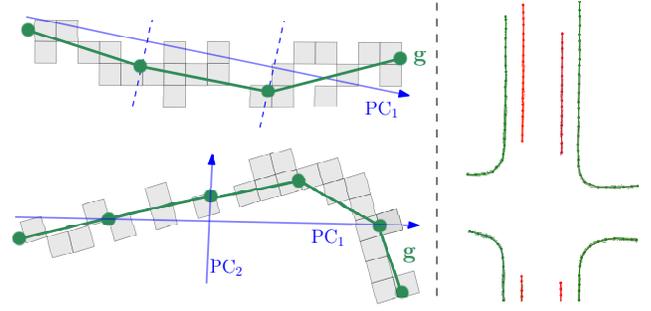
\begin{figure}[t]
\centering
\includegraphics[width=0.64\linewidth]{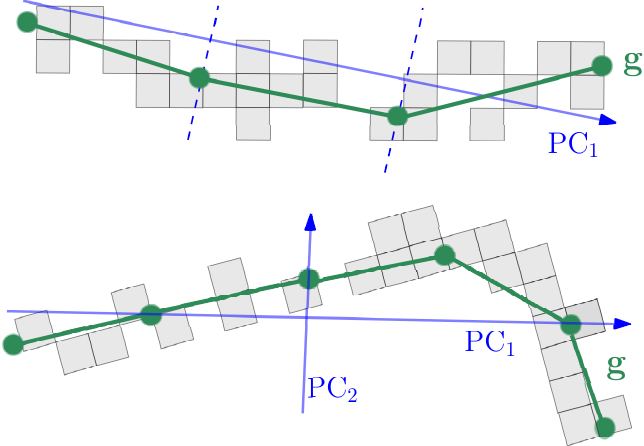}
\begin{tikzpicture}
    \draw[dashed] (0,0) -- (0,-4cm);
\end{tikzpicture}
\includegraphics[width=0.32\linewidth]{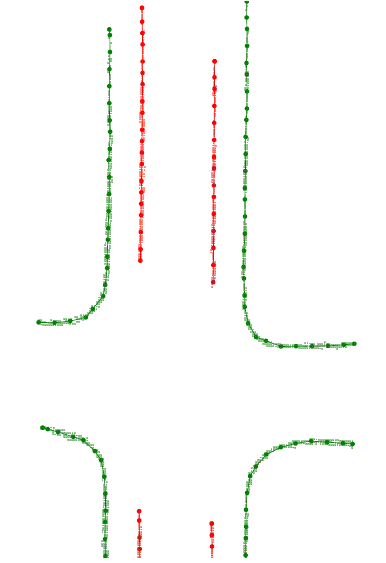}
\caption{
The voxels are divided into groups according to the principal components (\textcolor{blue}{blue}), and polylines (\textcolor{ForestGreen}{green}) are estimated based on the grouping.
\textbf{Top-left}: The voxels are distributed mainly along $\mathrm{PC}_1$ and can be effectively grouped based on the projections onto $\mathrm{PC}_1$.
\textbf{Bottom-left}: When $\mathrm{PC}_2$ is significant, the voxels are divided into the quadrants formed by $\mathrm{PC}_{1/2}$, and each quadrant is processed similarly as in the top-left case. \textbf{Right}: In a scene of the Argoverse2 dataset~\cite{Argoverse2}, lanelines (\textcolor{Red}{red}) corresponds to the left-top case, while roadedges (\textcolor{ForestGreen}{green}) corresponds to the left-bottom case.
}
% \vspace{-10pt}
\label{fig_geo_param_est}
\end{figure}

To make the map of reliable semantic voxels (see Fig.~\ref{fig_framework}) useful for downstream modules, we further cluster the semantic voxels to identify individual road markings and estimate their geometric parameters.
The process, referred to as instance mapping, is also done incrementally.

Formally, at each time $t$, we have a collection of instances for different types of road markings $\mathcal{I}^{(t)}=\bigcup_{l\in\mathcal{L}} \mathcal{I}^{(t)}_{l}$, and with a slightly abuse of notation,
\begin{align}
\mathcal{I}_{l}^{(t)} = \{I^l_i\}_{i=1}^{N_l^{(t)}},\;
I_i^l = (\{v_j\}^{M^{l}_{i}}_{j=1}, \mathbf{g}_{i}^{l})
\label{eq_instances_l}
\end{align}
where $\mathcal{I}_{l}^{(t)}$ denotes all the instances of type $l$. Each instance $I_{i}^{l}$ consists of a set of voxels and a parameterized geometric model $\mathbf{g}_{i}^{l}$. Initially, $\mathcal{I}^{(0)}$ is empty, and instances are built incrementally from $\mathcal{V}^{(t)}$ \eqref{eq_new_mature_vox}.

\begin{figure*}[t]
    {
        \centering
        \includegraphics[width=0.99\linewidth,trim={0 0.2cm 0 0},clip]{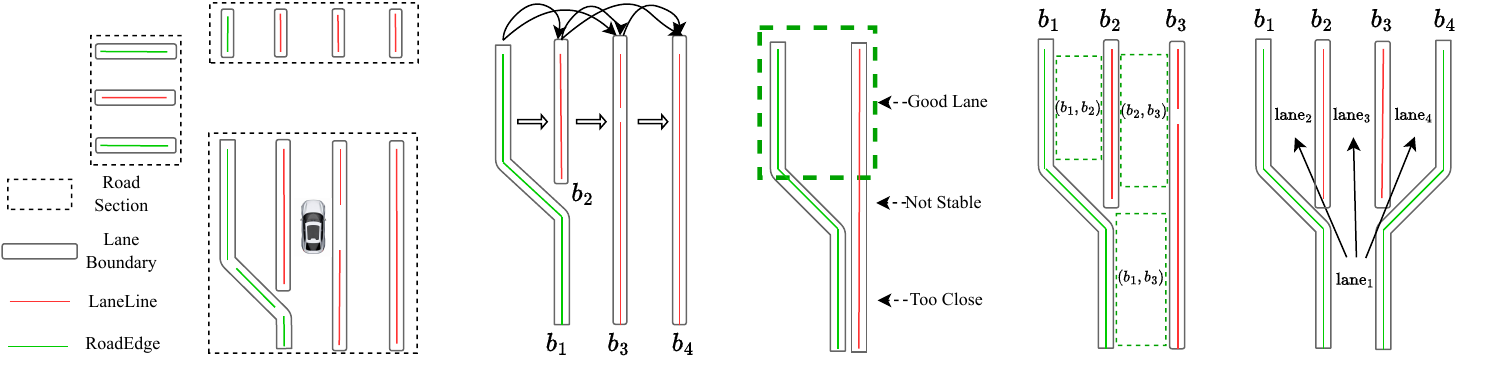}
    }\\
    \begin{minipage}{0.35\textwidth}
    \subcaption{\label{fig_road_sec}}
    \end{minipage}
    \begin{minipage}{0.12\textwidth}
    \subcaption{\label{fig_lanebound_sort}}
    \end{minipage}
    \begin{minipage}{0.17\textwidth}
    \subcaption{\label{fig_lane_builder}}
    \end{minipage}
    \begin{minipage}{0.2\textwidth}
    \subcaption{\label{fig_multi_lane}}
    \end{minipage}
    \begin{minipage}{0.08\textwidth}
    \subcaption{\label{fig_lane_linkage}}
    \end{minipage}
    \vspace{-1pt}
    \caption{
    Generating lanes and linkages from road marking instances.
    \textbf{(a)} Laneline and roadedge instances are grouped to continuous lane boundaries, and the lane boundaries are grouped into road sections.
    \textbf{(b)}
    Boundary pairs for candidate lanes are selected from the sorted (left-to-right) lane boundaries.
    \textbf{(c)}
    The distance between a pair of boundaries is examined to determine the valid range for a possible lane.
    \textbf{(d)}
    Following (b) and (c), correct lanes (green dashed rectangles) can be generated for situations where the number of lanes changes.
    \textbf{(e)}
    The linkages from $\text{lane}_1$ to $\text{lane}_{2/4}$ are determined by shared boundaries ($b_1$ and $b_4$ respectively). The linkage from $\text{lane}_1$ to $\text{lane}_3$ is established based on their geometric relation.
    }
    \label{fig_lane1}
% \vspace{-10pt}
\end{figure*}

\subsubsection{Incremental Clustering}
The newly available reliable semantic voxels $\mathcal{V}^{(t)} = \{v^{*}_{k}\}_{k=1}^{K}$ are used to update the existing semantic instances $\mathcal{I}^{(t-1)}$ or create new ones.
For each voxel $v^{*}_k$, we first select all the instances $\mathcal{I}^{(t-1)}_l$ that are of the same road marking type.
Then, for each instance $I^{l}_{i}$ in $\mathcal{I}^{(t-1)}_l$, we use the co-observation matrix $\mathcal{A}^{(t)}$ defined in \eqref{eq_co_obs_mat} to determine whether $v^{*}_k$ belongs to that instance. 
Specifically, for each voxel $v_j$ in $I_{i}^{l}$, we calculate the probability of the voxel and the candidate voxel $v^{*}_k$ belonging to the same instance as $p_{j} = \max(\frac{\mathcal{A}(v_j, v^{*}_k)}{n_j}, \frac{\mathcal{A}(v_j, v^{*}_k)}{n^{*}_k})$, where $p_{j}$ ranges from 0 to 1. We then count the voxels with large enough probability as $h_k = |\{ v_j \in I^{l}_{i} \ | \ p_{j} > \beta_{p}\}|$, and $v^{*}_k$ is considered belonging to the instance and added when $h_k > \beta_{n} \quad \text{or} \quad h_k / M_i > \beta_{r}$, where $\beta_n$ and $\beta_r$ are the minimum number and ratio of the voxels that are likely to belong to the same instance respectively.
If the voxel does not match any of the instances, a new instance is created.

The co-observation matrix retains the instance-level information in road marking detection, in addition to the geometric information accumulated in each voxel.
Clustering with the co-observation matrix has the advantage of not requiring complicated geometric thresholds and works quite robustly in our experience.

\subsubsection{Polyline Model Estimation}
We use a polyline as the parametric model for a road marking, and the parameters of the geometric model in \eqref{eq_instances_l} become $\mathbf{g}_{i} = \{\mathbf{y}_k\}_{k=1}^{H_i}$.
Polylines are simple yet effective models to represent the shapes of the road markings.
For example, stoplines can be compactly modeled as two-point polylines, and curved lanelines/roadedges can easily be modeled with more points.

To estimate a polyline from the voxels in each road marking instance \eqref{eq_instances_l}, we essentially need to divide the voxels into groups, each of which can be fitted as a straight line with sufficiently low error.
We thus devise a division method based on principal component analysis (PCA), as illustrated in Fig.~\ref{fig_geo_param_est}.
First, the first and second principal components of the voxel centers, denoted as $\mathrm{PC}_1$ and $\mathrm{PC}_2$,  are computed, and the corresponding eigenvalues are $\lambda_1$ and $\lambda_2$. Then:
1) If $\lambda_2 / \lambda_1$ is below a given threshold, the voxels are considered to be primarily distributed along $\mathrm{PC}_1$. We can then divide $\mathrm{PC}_1$ into segments according to a predefined length , and the voxels are grouped based on which segment the projections of their centers on $\mathrm{PC}_1$ fall into.
2) Otherwise, the voxels are first divided into four quadrants according to both $\mathrm{PC}_1$ and $\mathrm{PC}_2$, and the voxels in each quadrant is processed the same as in the first case, with a smaller segment length to account for possible large curvature.
Once the voxel groups are determined, we can easily fit a line for each and connect the fitted lines as a complete polyline.

\subsection{Road Layout Generation}
\label{sec_road_layout}

Based on the road marking instances \eqref{eq_instances_l}, we can further generate detailed road layout, \ie lanes and their linkages, which can be directly consumed by PnC algorithms.
%Note that the following steps only need to deal with the parametric models of the instances and are thus very efficient.

\subsubsection{Lane Boundary Generation}
The boundaries of lanes can be either roadedges or lanelines.
Ideally, each instance in $\mathcal{I}_{\text{laneline}}$ and  $\mathcal{I}_{\text{roadedge}}$ should correspond to a complete laneline or roadedge in the real world.
However, due to imperfect detection (\eg occlusion, network performance), a physical laneline or roadedge may correspond to several instances in 
\eqref{eq_instances_l}.
Therefore, we first group the instances in $\mathcal{I}_{\text{laneline}}$ and $\mathcal{I}_{\text{roadedge}}$ into \textit{lane boundaries} respectively.
In particular, if two instances of the same type have close endpoints and the directions at these endpoints are similar, the two instances are grouped together to form a continuous lane boundary.

\subsubsection{Lane Boundary Grouping}
The lanelines and roadedges on actual roads are naturally divided into different groups by intersections, and only those within each group are possible to form lanes.
Based on this observation, we first group the lane boundaries into different \textit{road sections}, as shown in Fig.~\ref{fig_road_sec}.
In particular, we treat each lane boundary as a vertex in a graph.
Two vertices are connected, if the two lane boundaries are of a similar direction and mutually overlapping.
Formally, a lane boundary overlaps with another, if there exists at least one point on the former that can be projected to one segment from the polyline model of the latter.
Then we use depth-first search to find the connected components (\ie road sections).
%each of which corresponds to a road section.

\begin{figure*}[t]
    \centering
    \quad \quad \quad \quad \quad  \textbf{Surrounding Views}  \quad \quad \quad \quad \quad \quad \quad \quad \quad \quad \, \textbf{MapTR} \quad \quad \quad \quad \quad \textbf{Ours}  \quad \quad \quad \quad \quad \quad \, \textbf{GT} 
    \includegraphics[width=0.99\linewidth]{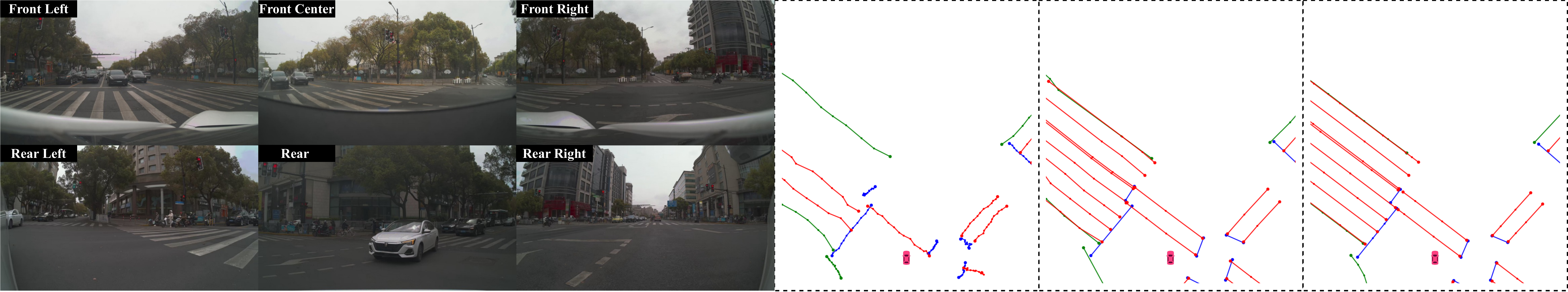}
    \includegraphics[width=0.99\linewidth]{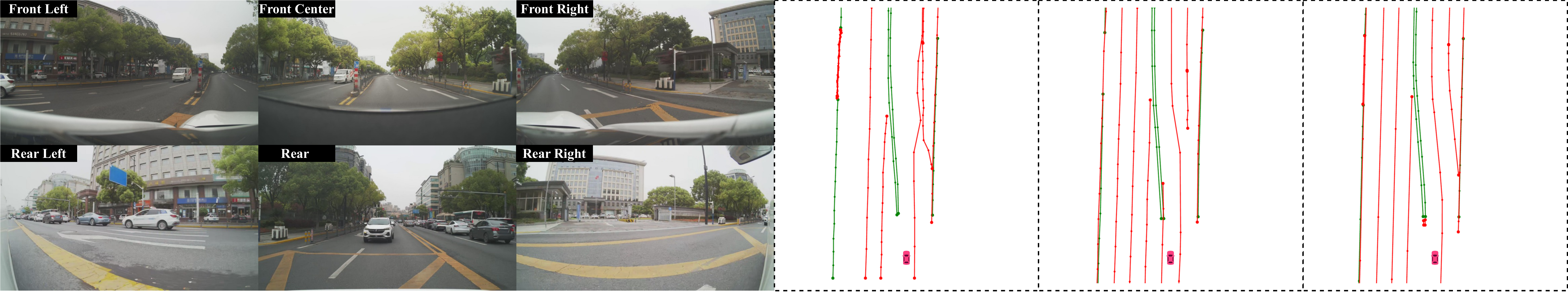}
    \caption{Qualitative results on the in-house dataset at intersections (top) and lane splitting/merging areas (bottom). The local map includes \textcolor{Red}{lanelines}, \textcolor{ForestGreen}{roadedges} and \textcolor{blue}{stoplines}.
    Our method is able to maintain accurate and complete road information (almost identical to the GT), while the single-frame detection from MapTR tends to be unstable in these situations (\eg due to the occlusion by the heavy traffic at intersections).
    }
    \vspace{-2pt}
    \label{fig:fig_self_qualitative}
\end{figure*}
\begin{figure*}[t]
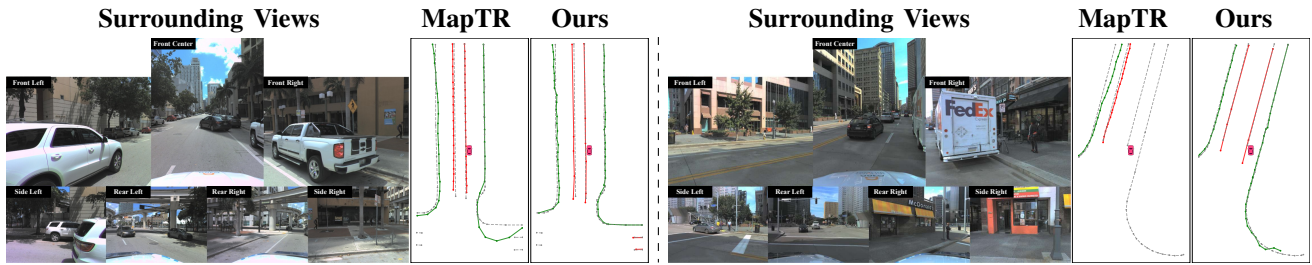

    \centering
    \begin{tabular}{>{\centering\arraybackslash}p{0.27\linewidth} >{\centering\arraybackslash}p{0.05\linewidth} >{\centering\arraybackslash}p{0.09\linewidth}>{\centering\arraybackslash}p{0.28\linewidth} >{\centering\arraybackslash}p{0.06\linewidth} >{\centering\arraybackslash}p{0.08\linewidth}}
        \textbf{Surrounding Views} & \textbf{MapTR} & \textbf{Ours} & \textbf{Surrounding Views} & \textbf{MapTR} & \textbf{Ours}
    \end{tabular}
    \\
    {
        \centering
        \includegraphics[width=0.48\linewidth]{fig/figures/fig_argo/SLM-argo3.pdf}
        \begin{tikzpicture}
            \draw[dashed] (0,0) -- (0,-3cm);
        \end{tikzpicture}
        \includegraphics[width=0.48\linewidth]{fig/figures/fig_argo/SLM-argo2.pdf}
    }
    \caption{
    Qualitative results on Argoverse2. Same color code as in Fig.~\ref{fig:fig_self_qualitative} is used, and the groundtruth is overlaid as dashed lines.
    Our method outputs more geometrically accurate road markings (note the better alignment between the groundtruth and our output) and are robust to the misdetection of the network.
    }
    % \vspace{-10pt}
    \label{fig:fig_self_qualitative_argo}
\end{figure*}

\subsubsection{Lane Generation}
To generate lanes within each road section, we design a lane generation process that are effectively for common lane layouts.
First, we sort the lane boundaries in the left-to-right order according to their relative position as illustrated in Fig.~\ref{fig_lanebound_sort}.
The sorted lane boundaries are denoted as $\mathcal{B}=\{b_i\}_{i=1}^{N_{\mathcal{B}}}$.
Then, for each lane boundary $b_i$, we select the two subsequent boundaries in $\mathcal{B}$ and form the boundary pairs for potential lanes, namely $(b_i, b_{i+1})$ and $(b_i, b_{i+2})$.
Finally, lanes are generated from the boundary pairs if certain conditions are satisfied.
In particular, for a pair $(b_\text{left}, b_\text{right})$, we sample points on $b_\text{right}$ and calculate the distances of the points to $b_\text{left}$.
A lane can be generated if, as shown in Fig.~\ref{fig_lane_builder}, there exists a range on $b_\text{left}$ where the distances within the range do not vary much and are within a preset distance interval (according the national standard).
A successful generated lane is thus represented by two boundaries and a range on the left one.
An illustration of the generated lanes for situations where the number of lanes changes along the road direction is shown in Fig.~\ref{fig_multi_lane}.

\subsubsection{
Non-Intersection Lane Linkage Generation
}
An lane linkage indicates whether a lane is the successor of another and is important for the planning module.
We primarily use two cues, as illustrated in Fig.~\ref{fig_lane_linkage}, to establish the linkages among the generated lanes in the non-intersection scenarios.
First, if two lanes share the same left or right boundary, a linkage will be created between the two.
Second, the geometric information of the lanes (\eg whether the end of one lane aligns well with the beginning of another) are used to create possible linkages as well.

\subsection{Discussion}
Our method discretizes and fuses the vectorized network detection into a voxel map and then re-creates the vectorized instances from the voxels.
While such a process may seem redundant, we would like to highlight that it removes the need to track the detected road markings, which have little temporal consistency guarantee.
Moreover, the voxel map along with the co-observation matrix allows for the accumulation of long-term temporal information efficiently, which enables our system to extract consistent and accurate instances, as will be shown in the next section.

\section{Experiments}

We evaluated the proposed method on both public and in-house data.
Variants of MapTR are used as the detection networks.
We present both quantitative and qualitative results to demonstrate the superior stability and accuracy compared with raw network output.
We also show that our approach has been successfully used in a mapless autonomous driving system in complicate urban environments.

\subsection{Datasets}

\subsubsection{In-house dataset}
The in-house dataset was recorded in typical urban areas and includes multiple trips by the same vehicle on different dates.
Each session lasted approximately 8 minutes and featured data from six synchronously triggered cameras at \SI{10}{\hertz}, alongside \SI{100}{\hertz} INS data.
For quantitative evaluation, we utilized a HD map as the groundtruth (GT).
We trained our own variant of MapTR using private training datasets.

\subsubsection{Argoverse2 dataset}
The validation dataset of Argoverse2~\cite{Argoverse2} includes 150 scenes, and each scene lasts around 15 seconds and contains \SI{10}{\hertz} RGB images from seven cameras. 
The validation dataset is accompanied by a 3D HD map, which we used as the groundtruth.
For road marking detection, we used the publicly available MapTRv2 \cite{liao2023maptrv2} model,
% \footnote{\url{https://github.com/hustvl/MapTR}}
which is one of the top performing models on Argoverse2.
%
%The model uses ResNet-50 as the backbone and was trained for 6 epochs.
%
Since the pretrained model does not include stoplines, our analysis focused on lanelines and roadedges only for Argoverse2.

\begin{table}[t]
\centering
\renewcommand{\arraystretch}{1.3}
\caption{Quantitative Comparison in the in-house Dataset}
\label{tab:comparison1}
\resizebox{\linewidth}{!}{
\begin{tabular}{c|cccccc}
\hline
\textbf{Category} & \textbf{Type} & \textbf{Method} & \textbf{Precision[\%] $\uparrow$} & \textbf{Recall[\%] $\uparrow$} & \textbf{F1[\%] $\uparrow$} & \textbf{ACD[m] $\downarrow$} \\
\hline
 & \multirow{2}{*}{Laneline} & MapTR & 77.23 & 74.56 & 75.87 & 0.164\\
& & Ours & \textbf{83.82} & \textbf{74.80} & \textbf{79.06} & \textbf{0.120}\\
\cline{2-7}
& \multirow{2}{*}{Stopline} & MapTR & 71.88 & 76.84 & 74.28 & 0.167\\
\multirow{1}{*}{Road} & & Ours & \textbf{73.36} & \textbf{77.55} & \textbf{75.40} & \textbf{0.146}\\
\cline{2-7}
\multirow{1}{*}{Markings} & \multirow{2}{*}{Roadedge} & MapTR & 61.78 & 70.32 & 65.78 & 0.148\\
& & Ours & \textbf{84.87} & \textbf{75.99} & \textbf{80.19} & \textbf{0.136}\\
\cline{2-7}
& \multirow{2}{*}{Total} & MapTR & 71.59 & 73.49 & 72.52 & 0.158\\
& & Ours & \textbf{82.81} & \textbf{75.41} & \textbf{78.93} & \textbf{0.128}\\
\cline{1-7}
% \cline{1-7}
\multirow{1}{*}{Road Layout} & \multirow{1}{*}{Lane} & Ours & 64.87 & 62.38 & 63.60 & 0.145 \\
\hline
\end{tabular}
}
% \vspace{-10pt}
\end{table}

\subsection{Map Quality Metrics}
\label{map_quality_metrics}

We use precision, recall, and F1 score as instance-level metrics. 
%First, the predicted vector lines and the groundtruth (GT) vector lines are sampled at 0.1m intervals to generate two dense point clouds.
To determine whether a predicted line matches a GT line, we first sample points from the two at the interval of \SI{0.1}{\m}.
A sampled point on the predicted line is considered successfully matched to the GT if its smallest distance to the sampled GT points is less than \SI{0.5}{\m}.
The predicted line is considered a true positive (TP) if the number of successfully matched points in it exceeds 75\% of the samples in the GT line. Additionally, each GT line can only be matched to one predicted line.
Then the precision, recall and F1 score can be computed according to common definitions.

Average Chamfer distance (ACD) is used as the point-level metric to indicate the geometric accuracy.
The Chamfer distance (CD) of a predicted line is computed if it is matched successfully to a GT line.
It is the average of the distances between the matched sample points and the corresponding GT points.
ACD is the average of the Chamfer distances of all recognized TP pairs.

\subsection{Mapping Results}

The proposed system is implemented to generate a local map. Only the map (voxels, instances and road layout) within a certain range around the vehicle is kept, and the information out of the range is cleared at every frame.
We set the mapping range according to the range of the network output.
The range is $[\SI{-15.0}{m}, \SI{15.0}{m}]$ on the side and $[\SI{-5.0}{\m}, \SI{30.0}{\m}]$ along the forward direction for the in-house datasets.
The latter range is adjusted to $[\SI{-30.0}{m}, \SI{20.0}{m}]$ for Argoverse2.
The ranges are selected to be slightly shorter than the network detection range to account for the inherent delay in the temporal fusion/filtering process.
Most of the system parameters are kept the same for both datasets (see Fig.~\ref{fig:ablation_study}), except that $\alpha_n$ is set to 3 for Argoverse2.
%Most of the system parameters are kept the same for both datasets, such as a voxel size of $0.2$, $\beta_p = 0.6$, $\beta_n = 3$, $\beta_r = 0.7$, and $\alpha_r = 0$. The only exception is $\alpha_n$, which is set to 10 for the In-house dataset and 3 for the Argoverse2 dataset, based on the differing detection qualities observed in each dataset.

As shown in Table~\ref{tab:comparison1}, our method outperforms MapTR for all road marking types in every metric on our in-house dataset.
Overall, our method shows improvements in precision, recall, and F1 score by $11.22\%$, $1.92\%$, and $6.41\%$, respectively.
The large improvement in precision indicates that our method contains far fewer erroneous road markings. This is due to the fact that the voxel-based fusion is very effective in rejecting outliers.
In terms of recall, we suspect that the relatively small improvement is partly due to the property of the voxel voting strategy. It needs to accumulate a certain number of detection of a road marking to generate the corresponding instance output, which results in lower recall in the first few frames.
Nevertheless, by keeping a persistent map, our method is less affected by occlusion, where the single-frame detection from MapTR may miss previous detected road markings.
The output from our method is also more geometrically accurate than the raw network output in terms of ACD.
We can observe a similar trend in the quantitative results on Argoverse2, as shown in Table~\ref{tab:comparison_av2_02_4_quad_div}.
Our method consistently improves the network output in terms of different metrics.
This also demonstrates that the effectiveness of our method is agnostic to the detection performance to a certain extent.
The lane generation process relies on accurate instances. For completeness, we also evaluated the centerlines of the generated lanes using the same protocol on the in-house dataset, as shown in Table.~\ref{tab:comparison1}.

\begin{table}[t]
\centering
\renewcommand{\arraystretch}{1.3}
\caption{Quantitative Comparison in Argoverse2}
\label{tab:comparison_av2_02_4_quad_div}
\resizebox{\linewidth}{!}{
\begin{tabular}{ccccccc}
\hline
\textbf{Type} & \textbf{Method} & \textbf{Precision[\%] $\uparrow$} & \textbf{Recall[\%] $\uparrow$} & \textbf{F1[\%] $\uparrow$} & \textbf{ACD[m] $\downarrow$} \\
\hline
\multirow{2}{*}{Laneline} & MapTR & 71.90& 66.24& 68.96& 0.134\\
& Ours & \textbf{77.40}& \textbf{74.50}& \textbf{75.91}& \textbf{0.123}\\
\hline
\multirow{2}{*}{Roadedge} & MapTR & 54.32& 51.41& 52.82& 0.176\\
& Ours & \textbf{56.48}& \textbf{53.01}& \textbf{54.52}& \textbf{0.167}\\
\hline
\multirow{2}{*}{Total} & MapTR & 62.00& 58.30& 60.09& 0.156\\
& Ours & \textbf{65.11}& \textbf{62.49}& \textbf{63.77}& \textbf{0.147}\\
\hline
\end{tabular}
}
% \vspace{-10pt}
\end{table}

 In Fig.~\ref{fig:fig_self_qualitative}, we highlight the performance of our method in circumstances where the network detection tends to deteriorate on the in-house dataset.
 We find that, at intersections and in lane merging/splitting areas, the network was more likely to misdetect certain markings or output geometrically wrong results compared with straight roads.
 These complicated road layouts, however, are more difficult for PnC algorithms than straight roads, and thus reliable road information is even more critical.
 In contrast, our method was able to maintain more accurate and complete road marking information in spite of the instability of the single frame detection.
 We observe similar improvements on Argoverse2 as well, as shown in Fig.~\ref{fig:fig_self_qualitative_argo}.

%\revision{To evaluate the quantitative results of the lanes generated by our method on the in-house dataset, we estimate the lane centerline using the left and right lane boundaries, then apply the same evaluation metrics described in \ref{map_quality_metrics} based on this estimated centerline. The table~\ref{tab:comparison1} presents the performance of our method in lane prediction tasks. Our method achieves a precision of $64.87\%$ and a recall of $62.38\%,$ resulting in an F1 score of $63.60\%$. Compared to the instance-level results, our lane prediction accuracy is lower. This is because our lane generation relies on instances, and the quality and availability of these instances directly affect the accuracy of lane prediction.}

\subsection{Application: Urban Autonomous Driving}

The proposed method has been integrated in a complete autonomous driving system and tested on over \SI{100}{\km} urban roads.
% in three cities in China.
%
Overall, as shown in Fig.~\ref{fig_teaser}, our method was able to generate consistent and accurate road information in typical urban scenarios such as split, straight and intersection scenes. Importantly, the lanes and linkages proved to be an effective input for PnC algorithms for driving functions.
We encourage the reader to check the accompanying video for more qualitative results.
%An example of the test vehicle performing lane change maneuver based on the estimated lanes and linkages is illustrated in Fig.~\ref{fig_ex_lanejuction}.
%
Note that, in practice, the linkages cross major intersections were queried from a standard-definition (SD) map to supplement the generated map, since the other sides of large intersections are often not visible. Otherwise, the full autonomous driving system relies on no prior map.
The road marking types (\eg solid/dash) were estimated by matching the fused lane boundaries with the network detection.

\subsection{Ablation Study and Runtime}

\begin{figure}[t]
\centering
\includegraphics[width=0.99\linewidth,trim={0 0.35cm 0 0cm},clip]{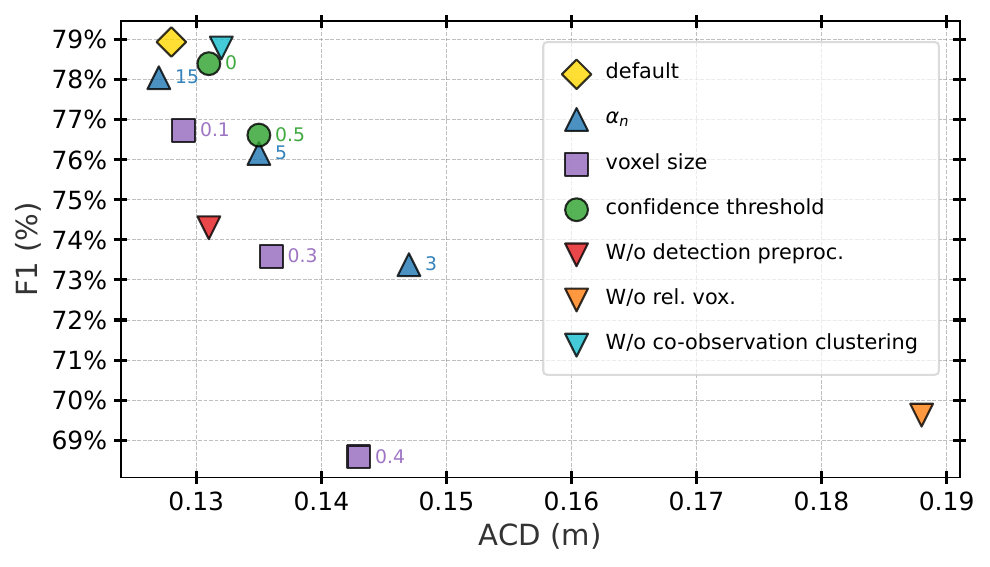}
\caption{
The ablation study on key parameters and modules. The default parameters are voxel size of \SI{0.2}{m}, confidence threshold of $0.3$, $\beta_p = 0.6$, $\beta_n = 3$, $\beta_r = 0.7$ and $\alpha_n=10$.
The values for the ablated parameters are marked on the right side of the corresponding marker.
``w/o detection preproc." indicates disabling the filters in Sec.~\ref{sec_detect_preproc}. ``w/o rel. vox." indicates $\alpha_n$ being set to $0$, and ``w/o co-observation clustering" indicates that in the clustering process, the nearest instance is selected out of all the instances that have been co-observed (instead of satisfying several thresholds).
}

% \vspace{-10pt}
\label{fig:ablation_study}
\end{figure}

\begin{figure}[t]
\centering
\includegraphics[width=\linewidth,trim={0 0.2cm 0 0},clip]{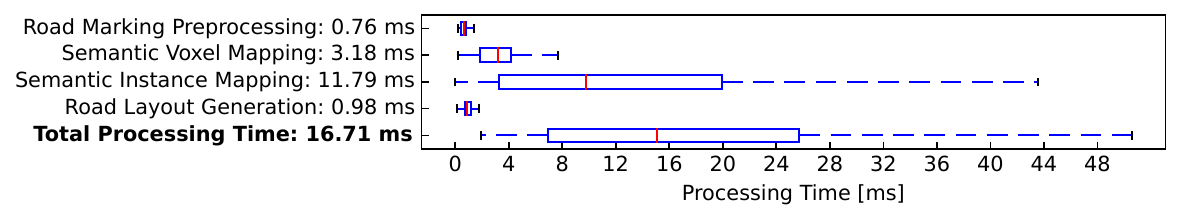}
\caption{Runtime on the NVIDIA DRIVE Orin platform.}
% \vspace{-14pt}
\label{fig_time}
\end{figure}

The runtime analysis of single-frame processing on the NVIDIA DRIVE Orin platform is shown in Fig.~\ref{fig_time}.
With an average processing time of just \SI{16.71}{\ms}, our system provides a large performance gain at a small computational cost and can easily meet the real-time requirement.
As for each sub-module, the semantic instance mapping is the most time consuming, as it needs to estimate polylines from a relatively large number of voxels.
The ablation studies for several key parameters and modules are shown in Fig.~\ref{fig:ablation_study}. Notably, some of the key designs, including voxel size and minimum detection times $\alpha_n$ for reliable voxels, turn out to be crucial for the overall performance.
%
%Note that our implementation is single-threaded, and estimating the geometric models for different instances can be further parallelized.
% \input{fig/fig_time}

\section{Conclusion}

In this letter, we have introduced a novel online road mapping system for autonomous vehicles that can generates highly consistent vectorized maps in real-time from image streams.
Thanks to the voxel-based fusion approach, we are able to efficiently fuse the noisy detection results from different times into consistent and accurate vectorized road marking instances.
Compared to single-frame online mapping methods, our system shows a large margin of improvement in terms of instance and geometric metrics and generates temporally more consistent results.
Additionally, from the improved road marking instances, our system further estimates lane and their linkages, thereby providing important information for downstream PnC modules.
The proposed system not only shows superior quantitative results but also has been validated extensively in a closed-loop autonomous driving system in real-world urban environments.
\revision{However, the system currently depends on SD maps to handle complex scenarios like intersections. While this approach ensures robustness, future work could focus on reducing reliance on external maps by directly inferring lane linkages from sensor data.}
% In future work, we plan to use widely available SD maps to improve the robustness of online map construction.
%, \eg to reject erroneous fusion results or complete missing road structure.
%
\revision{In addition, }we would also like to explore the possibility of retrieving geometrically meaningful uncertainty from the network and incorporate it in a probabilistic fusion process.

%%%%%%%%%%%%%%%%%%%%%%%%%%%%%%%%%%%%%%%%%%%%%%%%%%%%%%%%%%%%%%%%%%%%%%%%%%%%%%%%

\ifCLASSOPTIONcaptionsoff
  \newpage
\fi

\bibliographystyle{IEEEtran.bst}
\bibliography{reference}

\end{document}